\documentclass[letterpaper,10pt,conference]{ieeeconf}
\IEEEoverridecommandlockouts
\let\labelindent\relax
\usepackage[utf8]{inputenc} 
\usepackage[T1]{fontenc}    
\usepackage{graphicx} \graphicspath{ {figures/} }
\usepackage{amsmath,amssymb,mathabx,amsbsy,mathtools,etoolbox}
\usepackage{times}
\usepackage{graphicx}
\usepackage{amsmath,amssymb,mathbbol}
\usepackage{acronym}
\usepackage{enumitem}
\usepackage[breaklinks,colorlinks,linkcolor=black]{hyperref}
\usepackage{balance}
\usepackage{xspace}
\usepackage{setspace}
\usepackage[skip=3pt,font=small]{subcaption}
\usepackage[skip=3pt,font=small]{caption}
\usepackage[dvipsnames,svgnames,x11names]{xcolor}
\usepackage[capitalise,noabbrev,nameinlink]{cleveref}
\usepackage{booktabs,tabularx,colortbl,multirow,multicol,array,makecell}
\usepackage{dblfloatfix}
\usepackage[misc]{ifsym}
\usepackage{pifont}
\usepackage{cite}

\makeatletter
\DeclareRobustCommand\onedot{\futurelet\@let@token\@onedot}
\def\@onedot{\ifx\@let@token.\else.\null\fi\xspace}
\def\eg{\emph{e.g}\onedot} 

\def\ie{\emph{i.e}\onedot}

\def\wrt{w.r.t\onedot}

\makeatother

\makeatletter
\def\BState{\State\hskip-\ALG@thistlm}
\makeatother

\makeatletter
\renewcommand{\paragraph}{%
  \@startsection{paragraph}{4}%
  {\z@}{0ex \@plus 0ex \@minus 0ex}{-1em}%
  {\hskip\parindent\normalfont\normalsize\bfseries}%
}
\makeatother

\crefname{algorithm}{Alg.}{Algs.}
\Crefname{algocf}{Algorithm}{Algorithms}
\crefname{section}{Sec.}{Secs.}
\Crefname{section}{Section}{Sections}
\crefname{table}{Tab.}{Tabs.}
\Crefname{table}{Table}{Tables}
\crefname{figure}{Fig.}{Fig.}
\Crefname{figure}{Figure}{Figure}

\definecolor{gblue}{HTML}{4285F4}
\definecolor{gred}{HTML}{DB4437}
\definecolor{ggreen}{HTML}{0F9D58}

\definecolor{mygray}{gray}{.92}

\acrodef{dof}[DoF]{degree of freedom}
\acrodef{vkc}[VKC]{virtual kinematic chain}
\acrodef{tamp}[TAMP]{task and motion planning}
\acrodef{pddl}[PDDL]{planning domain definition language}
\acrodef{rrt}[RRT]{rapidly-exploring random tree}
\acrodef{ompl}[OMPL]{open motion planning library}
\acrodef{iws}[IWS]{iterated width search}
\acrodef{bfs}[BFS]{breadth first search}
\acrodef{dfs}[DFS]{depth first search}
\acrodef{ai}[AI]{artificial intelligence}
\acrodef{spt}[SPT]{scene parse tree}
\acrodef{com}[CoM]{center of mass}
\acrodef{mcmc}[MCMC]{Markov chain Monte Carlo}
\acrodef{ged}[GED]{graph editing distance}
\acrodef{sdf}[SDF]{signed distance field}
\acrodef{cma}[CMA-ES]{covariance matrix adaptation evolution strategy}
\acrodef{asa}[ASA]{adaptive simulated annealing}
\acrodef{gmm}[GMM]{Gaussian mixture models}

\title{\LARGE \bf Rearrange Indoor Scenes for Human-Robot Co-Activity}

\author{Weiqi Wang$^{1,\star}$\quad{}Zihang Zhao$^{2,3,\star}$\quad{}Ziyuan Jiao$^{1,2,\star}$\quad{}Yixin Zhu$^{4,\dagger}$\quad{}Song-Chun Zhu$^{2,4}$\quad{}Hangxin Liu$^{2,\dagger}$\quad{}
\vspace{6pt}%
\\\url{https://sites.google.com/view/coactivity}%
\vspace{3pt}%
\thanks{$^{\star}$ W. Wang, Z. Zhao, and Z. Jiao contributed equally to this work.
$^\dagger$ Corresponding authors. Emails:
\texttt{yixin.zhu@pku.edu.cn},
\texttt{liuhx@bigai.ai}.
}%
\thanks{
$^{1}$ UCLA Center for Vision, Cognition, Learning, and Autonomy (VCLA)
$^{2}$ National Key Laboratory of General Artificial Intelligence, Beijing Institute for General Artificial Intelligence (BIGAI)
$^{3}$ College of Engineering, Peking University
$^{4}$ Institute for Artificial Intelligence, Peking University%
}%
}

\begin{document}

\maketitle
\thispagestyle{empty}
\pagestyle{empty}

\begin{abstract}
We present an optimization-based framework for rearranging indoor furniture to accommodate human-robot co-activities better. The rearrangement aims to afford sufficient accessible space for robot activities without compromising everyday human activities. To retain human activities, our algorithm preserves the functional relations among furniture by integrating spatial and semantic co-occurrence extracted from SUNCG and ConceptNet, respectively. By defining the robot's accessible space by the amount of open space it can traverse and the number of objects it can reach, we formulate the rearrangement for human-robot co-activity as an optimization problem, solved by \ac{asa} and \ac{cma}. Our experiments on the SUNCG dataset quantitatively show that rearranged scenes provide a robot with 14\% more accessible space and 30\% more objects to interact with on average. The quality of the rearranged scenes is qualitatively validated by a human study, indicating the efficacy of the proposed strategy.
\end{abstract}

\vspace{3pt}
\section{Introduction}

Service robots are gaining popularity in domestic settings, where they are expected to perform various complex household tasks. Typically, indoor scenes are designed and organized in accordance with human needs, often too confined and clustered for conventional service robots to navigate and interact. To tackle these challenges, researchers have developed several effective planning algorithms, designed to (i) retrieve objects in confined and cluttered spaces~\cite{berenson2008optimization,cheong2020relocate,han2020towards}, (ii) coordinate whole-body motions for articulate objects~\cite{shankar2016kinematics,bodily2017motion,chitta2010planning}, (iii) coordinate foot-arm via virtual mechanisms for sequential manipulation tasks~\cite{jiao2021virtual,jiao2021efficient}, and (iv) integrate robot perception and task planning for household environments~\cite{han2021reconstructing,jiao2022sequential,han2022scene}. Nevertheless, indoor scenes designed purely for humans may fundamentally prohibit robot activities due to their different morphology and movement patterns; robots with bulky embodiments have difficulty in imitating human motions and require larger open spaces.

\begin{figure}[t!]
    \centering
    \begin{subfigure}[t]{0.5\linewidth}
        \centering
        \includegraphics[width=\linewidth]{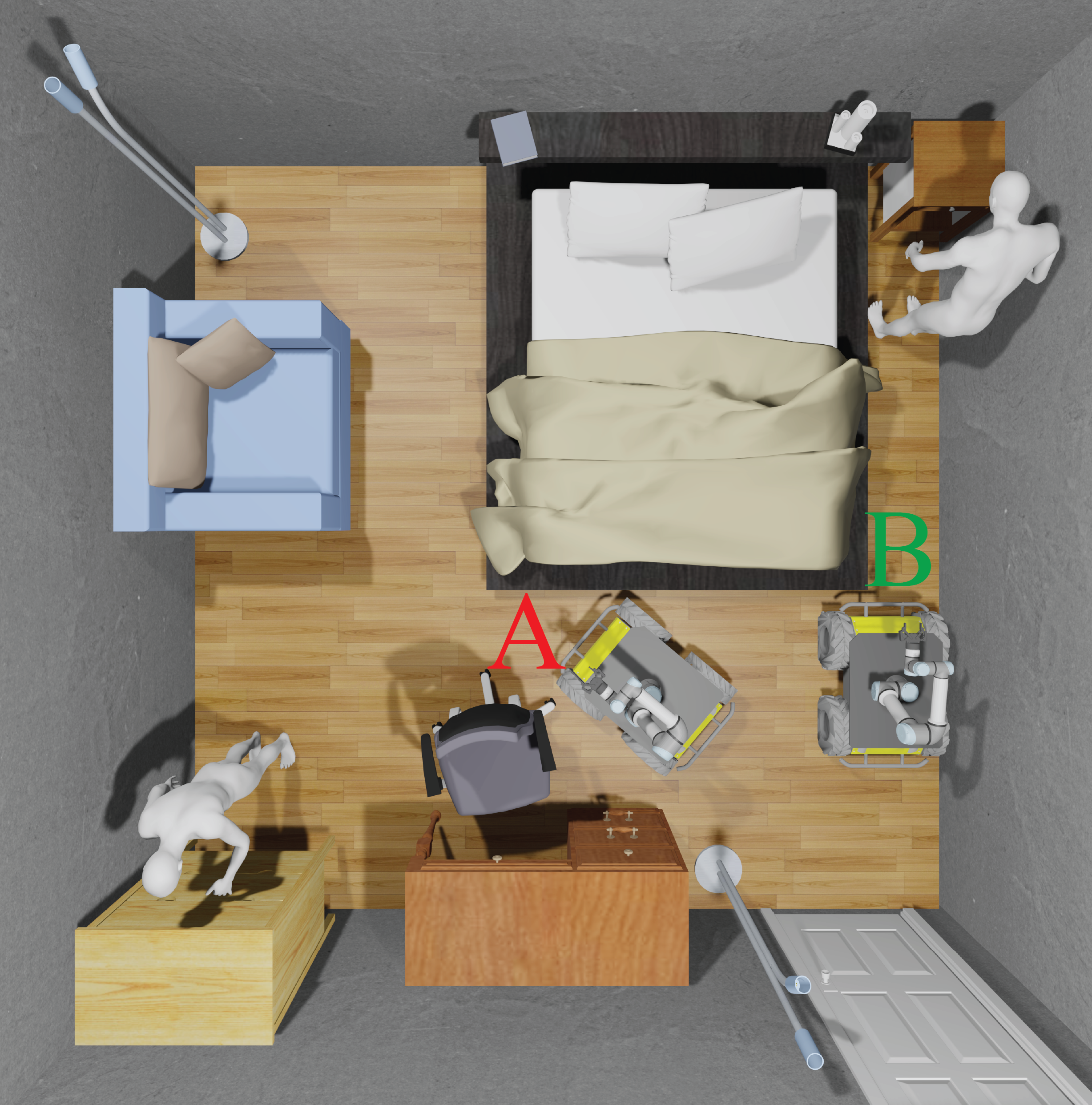}%
        \caption{human activity only}
        \label{fig:motivation_human}
    \end{subfigure}%
    \begin{subfigure}[t]{0.5\linewidth}
        \centering 
        \includegraphics[width=\linewidth]{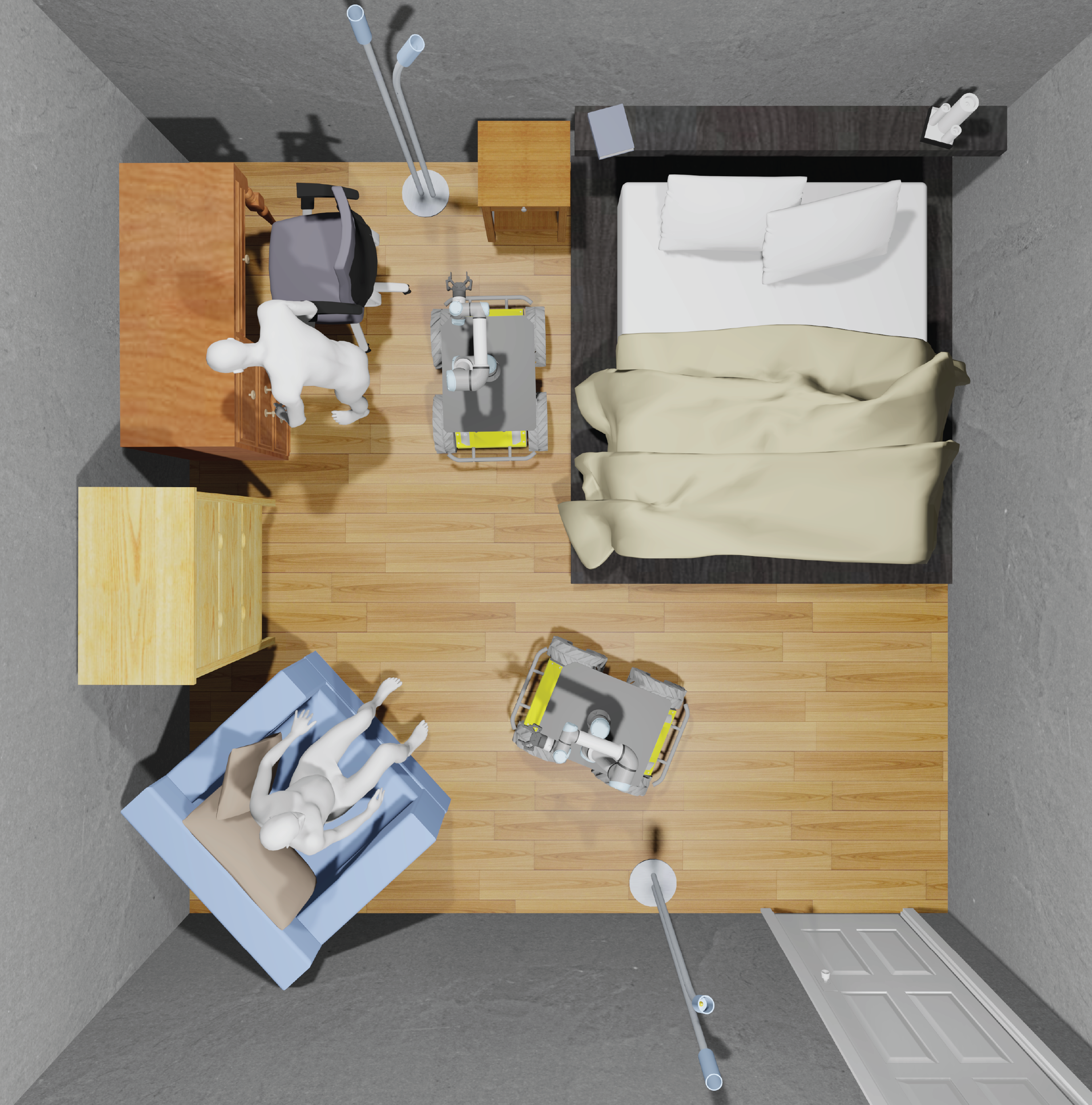}
        \caption{human-robot co-activity}
        \label{fig:motivation_codesign}
    \end{subfigure}%
    \caption{\textbf{Rearrange indoor scenes for human-robot co-activity.} (a) A person may pass through the narrow passages at {\color{OrangeRed} A} and {\color{LimeGreen}{B}} for daily activities, whereas a robot cannot due to its larger footprint. As a result, the robot's activities are limited in household environments designed purely for human activities. (b) A rearranged scene, optimized for human-robot co-activity, provides sufficient open space for robot activities while preserving human preferences.}
    \label{fig:motivation}
\end{figure}

\cref{fig:motivation_human} illustrates the above problem. In this example, a person may easily approach and interact with the nightstand, whereas a robot cannot due to its bulky mobile base that is larger than the space between the bed and the wall. Similarly, a robot cannot reach the bookshelf due to the narrow passageway between the chair and the bed. These prevalent restrictions of service robots significantly restrict the robot's capabilities in household environments. Notably, better planning algorithms cannot solve this problem; the room layout must be optimized for human-robot co-activity.

We argue that the essence of scene rearrangement for human-robot co-activity is to preserve human preference for indoor activities while affording more robot activities in terms of accessible space and interactions with objects. This dual objective requires simultaneously modeling both \textbf{human preference} and \textbf{robot preference} in the indoor scenes.

\textbf{Scene synthesis} has primarily modeled human preference and automatically generates scene layout from scratch subjects to constraints. Classic methods exploit simple heuristics to construct these constraints, such as interior design guidelines~\cite{merrell2011interactive} or user-provided positive examples~\cite{fisher2012example}. Modern treatments adopt learning-based approach, such as learning object-object relations~\cite{yu2011make,kermani2016learning}, modeling human activities with objects~\cite{qi2018human,jiang2018configurable}, mining topological relationships among object groups~\cite{keshavarzi2020scenegen}, and capturing latent information via generative models~\cite{ritchie2019fast,zhang2020deep,nauata2021house}. In particular, the common assumption in scene synthesis is that large datasets capture the statistics (\ie, scene layouts) for downstream tasks. This assumption no longer holds for human-robot co-activity as existing datasets only capture human preference, lacking statistics to model their robot counterpart.

\textbf{Scene rearrangement} has modeled either human or robot preference individually, such as to (i) reduce the risk of patient falls~\cite{chaeibakhsh2021optimizing}, (ii) improve robot navigation efficiency~\cite{zhi2021designing}, (iii) boost workspace task performance~\cite{liang2019functional}, and (iv) promote collaboration~\cite{zhang2021joint}. The lack of joint modeling of both human and robot preferences calls for alternative approaches.

To model \textbf{human preference}, we encode it implicitly by functional groups of objects (\ie, furniture)~\cite{zhao2011image,chen2019holistic++}. Intuitively, we place a nightstand beside a bed and a chair alongside a table. More broadly, objects within a functional group should be (re)arranged together, whereas their relative poses can vary within a smaller range. As such, a straightforward idea to model functional groups is based on spatial co-occurrence, \ie, the frequency with which two objects appear together and in close proximity~\cite{fisher2012example,yu2011make,wang2019planit,zhang2021fast}. Nonetheless, such a statistical perspective might be deceiving; the proximity of two objects does not necessarily suggest that they belong to the same functional group, especially in cluttered indoor scenes. In \cref{fig:motivation_human}, the chair and the bed, as well as the nightstand and the bed, are pretty close to each other. However, the bed and the chair are not in the same functional group. To overcome the ambiguity stemming from spatial co-occurrence, we employ ConceptNet~\cite{speer2017conceptnet}, a sizeable open-source knowledge graph database, to refine functional relations based on additional semantic co-occurrence of objects.

To model \textbf{robot preference}, we utilize \textit{accessible space}: the amount of open space a robot can traverse and the number of objects it can reach. Intuitively, a scene must afford sufficient open space for a robot to explore while performing given tasks. Objects must also be properly oriented for successful manipulation; for instance, a desk can be placed against a wall, but cabinets and drawers must face outside. To effectively encode and compute various possible interactions between a robot and a scene, we introduce a \ac{sdf} to represent (i) the scene's navigable area, given robot footprint and object placements, and (ii) the interaction affordance defined on the object boundary, akin to ``dark matters''~\cite{xie2017learning,zhu2020dark} that attract or repel possible interactions.

Computationally, we design an optimization framework of scene rearrangement for human-robot co-activity that takes as input the above robot and human preferences. \cref{fig:motivation_codesign} illustrates an exemplar result, in which robot and human preferences are co-optimized: (i) the robot functions more efficiently due to larger open space and potentially more interactions with objects, and (ii) the objects within functional groups remain close to satisfying human preference. In the experiment, we evaluate our method using the SUNCG dataset~\cite{song2017semantic}. Not only do the rearranged scene layouts afford an average of 30\% more robot activities and 14\% more open space, but also keep the \textit{Naturalness} based on a human study. 

Our \textbf{contributions} are threefold: (i) We develop a new method to capture human preferences via functional relations among objects by combining the spatial and semantic co-occurrence. (ii) We model robot preferences by its accessible space, represented by an \ac{sdf} for efficient computation. (iii) We devise an optimization-based framework to balance human and robot preferences when rearranging scene layouts.

The remainder of this paper is organized as follows. \cref{sec:preference} formally describes our modeling of human and robot preferences, subsequently formulated as an optimization framework presented in \cref{sec:optimization}. Experimental results are presented in \cref{sec:result}. We conclude the paper in \cref{sec:conclude}.

\begin{figure*}[t!]
    \centering
    \includegraphics[width=0.99\linewidth]{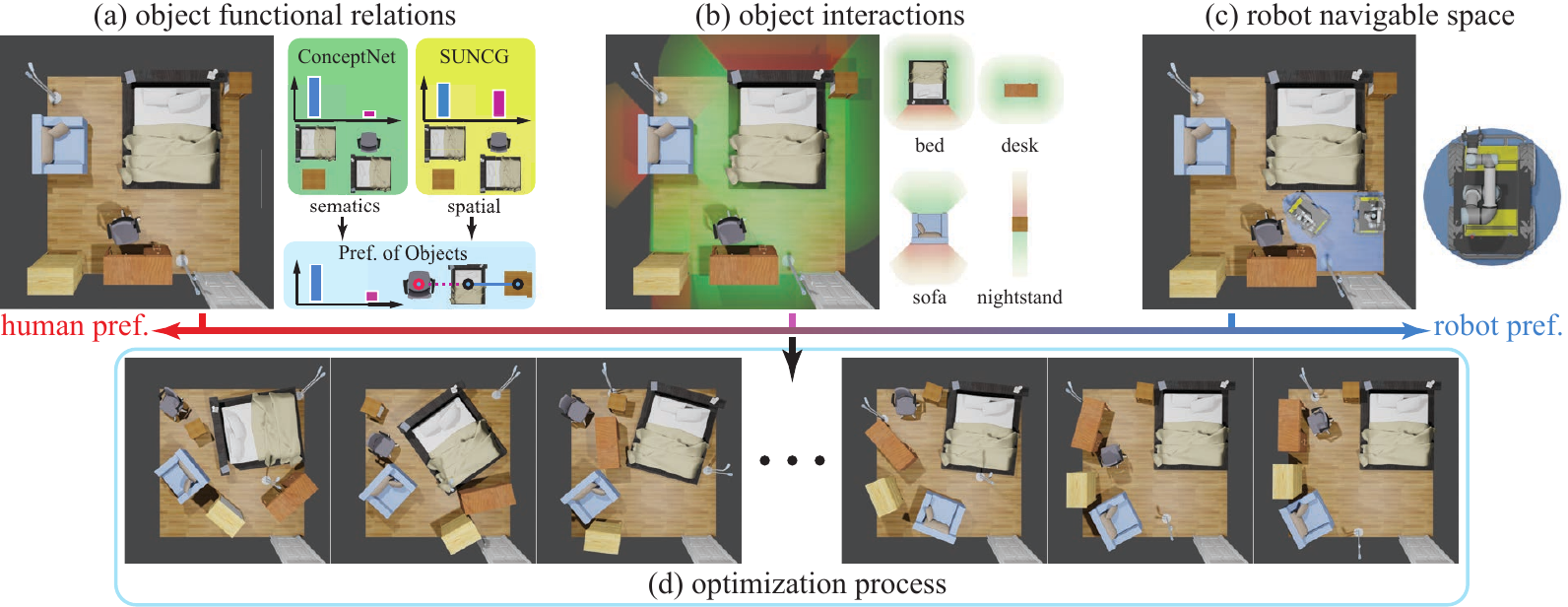}
    \caption{\textbf{Essential factors to rearrange scenes for human-robot co-activity.}
    (a) To preserve human preference encoded by object relations, we model object co-occurrence using semantic and spatial relations extracted from ConceptNet and SUNCG, respectively. (b) Robots can only interact with certain furniture from specific directions, such as approaching a cabinet or drawer from the front. Hence, we devise a pseudo-interaction function to indicate the desired furniture orientation. (c) After fitting an inflated circular base to the robot's footprint, the robot's accessible space in the scene (shaded blue) can be determined by an \ac{sdf}. (d) These three factors are formalized into an optimization framework to rearrange scenes for human-robot co-activity.}
    \label{fig:pipeline}
\end{figure*}

\section{Human and Robot Preferences}\label{sec:preference}

In this section, we describe how we model (i) human preference by combining objects' semantic and spatial co-occurrences to determine functional object groups and (ii) robot preference by its accessible space in the scene.

\subsection{Human Preference}

We model human preference implicitly by functional object groups. Discovering functional groups that emerge from objects in cluttered scenes is challenging due to the inherited ambiguity of objects' functional relations. Two nearby objects with close proximity do not necessarily indicate that they are functionally related (\eg, the example of the bed and chair shown in \cref{fig:pipeline}a), and objects with similar semantics (\eg, a desk and a coffee table) also do not imply that they are within a functional group. Hence, relying solely on spatial or semantics alone cannot correctly uncover the functional groups. In this work, we seek an integrated solution.

\paragraph*{Objective}

We employ a weighted scene graph $ \mathcal{G} = (V, S)$ to represent the relations among objects within a scene. Specifically, an object is represented by a node $v_i = \langle o_i, B_i, p_i \rangle\in V$ in the scene graph, including object label $o_i$, oriented 3D bounding box $B_i$, and the object planar pose $p_i$. Each edge $s_{i,j} = \langle v_i, v_j,w_{ij}\rangle \in S$ is a tuple that encodes the relation between two objects $v_i$ and $v_j$. The edge weight $w_{ij}$ indicates how likely $v_i$ and $v_j$ are within a functional group: 
\begin{equation}
    \small
    w_{ij}  = P(v_i,v_j|\mathcal{G}) = \frac{1}{Z} P_{sem}(o_i,o_j|\mathcal{G})P_{spa}(o_i,o_j|\mathcal{G}),
    \label{eqn:edge_weight}
\end{equation}
where $Z$ is a normalizing factor obtained by summing over all the products of the pairwise semantic/spatial probabilities:
\begin{equation}
    \small
    Z = \sum_{o_i,o_j,i\neq j} P_{sem}(o_i,o_j|\mathcal{G})P_{spa}(o_i,o_j|\mathcal{G}),
\end{equation}
and $P_{sem}(\cdot)$ and $P_{spa}(\cdot)$ are the probabilities reflecting the semantic and spatial correlations of two objects, respectively.

\paragraph*{Semantic Relation}

To obtain two object's semantic relation $P_{sem}(o_i,o_j|\mathcal{G})$, we utilize ConceptNet~\cite{speer2017conceptnet}, an open-source knowledge graph database that characterizes the strength of the semantic relation between two object labels $o_i$ and $o_j$ by $h_{ij}\in[0,1]$; a greater $h_{ij}$ suggests a stronger semantic relation. Unfortunately, ConceptNet would assign a large $h_{ij}$ to two synonyms (\eg, a chair and an armchair, a desk and a table), resulting in wrong functional relations between similar objects. To tackle this issue, when $o_i$ is synonymous with $o_j$, we replace $h_{ij}$ returned from ConceptNet by $h$ averaged over all other pairs of objects in the scene to indicate a neutral relation: 
\begin{equation}
    \small
    h_{ij}^* =
    \begin{cases}
        h_{ij},  & \neg~\texttt{IsA}(o_i, o_j) \\
        \mathbb{Avg}(\{h_{mn}|\neg~\texttt{IsA}(o_m, o_n)\}), & \text{otherwise}  
    \end{cases},
\end{equation}
where $\texttt{IsA}$ is an edge type, indicating $o_i$ is synonymous with $o_j$. Notably, we do not set $h$ between two synonyms to 0 to avoid rearranging these two objects apart.
Taken together, the semantic relation in \cref{eqn:edge_weight} is given by: 
\begin{equation}
    \small
    P_{sem}(o_i,o_j|\mathcal{G}) = \frac{h_{ij}^*}{\sum_{s\in S}h}.
    \label{eqn:human_semantic}
\end{equation}

\paragraph*{Spatial Relation}

To determine how likely objects $o_i$ and $o_j$ are related based on their spatial distance $d_{ij}$, we query their co-occurrence in the SUNCG dataset:
\begin{equation}
    \small
    P_{spa}(o_i,o_j|\mathcal{G}) \propto P_d(d_{ij}|o_i,o_j)P_{co}(o_i,o_j),
    \label{eqn:human_spatial}
\end{equation}
where $P_d(\cdot|o_i,o_j)$ is the distribution of relative poses between $o_i$ and $o_j$ in the SUNCG dataset. In practice, $P_d(\cdot|o_i,o_j)$ is biased if $o_i$ and $o_j$ rarely co-occur in the same scene. Hence, we add a term $P_{co}$ for bias correction:
\begin{equation}
    \small
    P_{co}(o_i,o_j) = \frac{N_{ij}} {\min \left( \sum_{k\in O} N_{jk}, \sum_{k\in O} N_{ik} \right)},
\end{equation}
where $N_{ij}$ is the number of co-occurrences of $o_i$ and $o_j$ in the dataset, and $O$ is the set of all semantic labels.

\subsection{Robot Preference}

Due to the existence of narrow passages and obstacles, a human-made environment is unfit for the functioning of a service robot. By rearranging the scene, we seek to expand the open space for robot activity. Specifically, we model the robot preference for a scene based on its accessible space $\mathcal{R}$, which has a closed boundary.
The accessible space of a robot consists of two components: (i) the open space it can traverse and (ii) the number of objects it can reach.

\paragraph*{Size of Open Space}

The open space that a service robot can traverse can be effectively computed by:
\begin{equation}
    \small
    f_\mathcal{R}(q) = f_\mathcal{B}(q) - r_{b},
    \label{eqn:reach}
\end{equation}
where $f_\mathcal{B}:\mathbb{R}^2\rightarrow\mathbb{R}$ is an \ac{sdf} that measures the shortest signed Euclidean distance from a query point $q$ to the bounding boxes of objects in the scene. When the robot is outside, on the boundary, or inside (\ie, invalid robot pose), the \ac{sdf}'s value is larger than, equal to, or less than zero, respectively. After imposing a circular base inflated with a radius of $r_{b}$ as a safety margin, $f_\mathcal{R}:\mathbb{R}^2\rightarrow\mathbb{R}$ denotes the entire open space of the robot.

\paragraph*{Number of Interacting Objects}

Although the size of the open space reveals a robot's capability to traverse the environment, it does not necessarily reflect a robot's ability to interact with objects. For instance, the cabinet door should not face the wall, as a robot cannot interact with it otherwise. \cref{fig:pipeline}b provides other examples; the green shaded areas around different objects indicate how a robot may approach and interact with these objects, whereas the red shaded areas imply the opposite. Computationally, we further introduce a pseudo-interaction function:
\begin{equation}
    \small
    f_{\mathcal{I}_i}(q) = \text{sgn}(q)\cdot \max(0, -\frac{f^+_{\mathcal{B}_i}(q)}{d^\text{max}} + 1), 
    \label{eqn:interaction}
\end{equation}
where $f^+_{\mathcal{B}_i}(q)$ is an \ac{sdf} that only returns a positive distance from a query point $q$ (\eg, a robot pose) to the bounding box of the object $v_i$, $d^\text{max}$ is the normalizing factor set to the maximum distance that the robot arm can reach, and $\text{sgn}(\cdot)$ assigns positive weights to the green shaded areas in \cref{fig:pipeline}b (\ie, the area where a robot would interact with the object) and negative to the red shaded areas.

\section{Optimization}\label{sec:optimization}

After building both human and robot preferences, we devise an optimization framework that balances these preferences and rearranges scenes accordingly. The resulting scene supports better human-robot co-activity.

\subsection{Objectives}

\paragraph*{Human Term}

Recall that human preference is modeled as a weighted scene graph $\mathcal{G}$ defined in \cref{eqn:edge_weight}. We first cluster the edges in $\mathcal{G}$ \wrt the edge weight by \ac{gmm}. Next, assuming that edges within the same functional group have large weights, we prune cluster edges that have small weights, resulting in a filtered scene graph $\mathcal{G}^\star$. As a result, each connected sub-graph $\mathcal{G}^\star$ is a functional group. Finally, the human term is defined as:
\begin{equation}
    \small
    H_{s_{i,j}} = 1 - \frac{P_d \left( d|o_i,o_j; s_{ij}^\star \in \mathcal{G}^\star \right)}{\sup||P_d \left( \cdot|o_i,o_j; s_{ij}^\star \in \mathcal{G}^\star \right)||},
    \label{eqn:human_loss}
\end{equation}
where $s_{ij}^\star$ is the edge connecting $o_i$ and $o_j$ in $\mathcal{G}^\star$.

\paragraph*{Robot Term}

The robot term is the combination of \cref{eqn:reach,eqn:interaction}:
\begin{equation}
    \small
    I_i = - \int_{q\in\mathcal{I}_i\cap\mathcal{R}} f_{\mathcal{I}_i}(q) + \alpha f_\mathcal{R}(q) dq,
    \label{eqn:robot_loss}
\end{equation}
where $\alpha$ is an empirically set balancing constant. A smaller $I_i$ is preferred for more open space for the robot.

\paragraph*{Objective}

Let $\psi=\{p_i|v_i\in V\}$ denote the scene layout. We formulate the problem of rearranging scenes as an optimization problem that minimizes the human and robot terms introduced in \cref{eqn:human_loss,eqn:robot_loss}.
\begin{equation}
    \begin{aligned}
        \min_{\psi} \quad{} & \sum_{s^\star \in \mathcal{G}^\star}{H_s} + \beta \sum_{v_i\in V} I_i,\\
        \textrm{s.t.} \quad{} & d(v_i,v_j) > 0,\quad{} \forall v_i,v_j\in V, i\neq j,
    \end{aligned}
    \label{eqn:objective}
\end{equation}
where $\beta$ is an empirically set balancing constant and the constraint $d$ forces the minimum distance between object pairs to be positive; \ie, each object pair is collision-free.

\subsection{Optimization}

We adopt the \acf{asa}~\cite{ingber1993adaptive} to solve the above optimization problem. \ac{asa} searches for a global minimum in an energy landscape defined by the objective function. Specifically, a candidate scene layout is sampled from a uniform distribution and is accepted based on the Metropolis criterion. \ac{asa} adjusts step size after several optimization iterations to maintain an approximately equal number of accepted and rejected samples for each variable. In practice, we find that \ac{asa} excels in escaping local minima but struggles to converge due to the enormous search space.

\paragraph*{Search Space}

We hierarchically optimize the layout to reduce the search space. The optimization is decomposed into two steps given a native human-centric scene. First, each functional group of the scene is treated as a sub-scene and is optimized independently \wrt \cref{eqn:objective}. Second, an optimized functional group is treated as a single object, and the scene layout is optimized over functional groups. \cref{fig:pipeline}d shows an optimization process with intermediate layouts. 

\paragraph*{Convergence}

We adopt \acf{cma} to expedite convergence. \ac{cma} is a derivative-free stochastic method for numerical optimization, which repeatedly applies the survival of the fittest process to its population and rapidly converges to a nearby local minimum. At the beginning of each iteration, \ac{cma} draws samples from a multivariate normal distribution: $\psi' \sim \mathcal{N}(m, \sigma^2 C)$, where $m$ is the weighted mean of the most promising layouts of previous samples, $\sigma$ is the overall standard deviation or step size, and $C$ is the estimated covariance matrix. At the end of each iteration, the algorithm updates these parameters according to the performance of the population, shifting the expected variance in the same direction as the estimated gradient.

\begin{figure*}[t!]
    \centering
    \includegraphics[width=\linewidth]{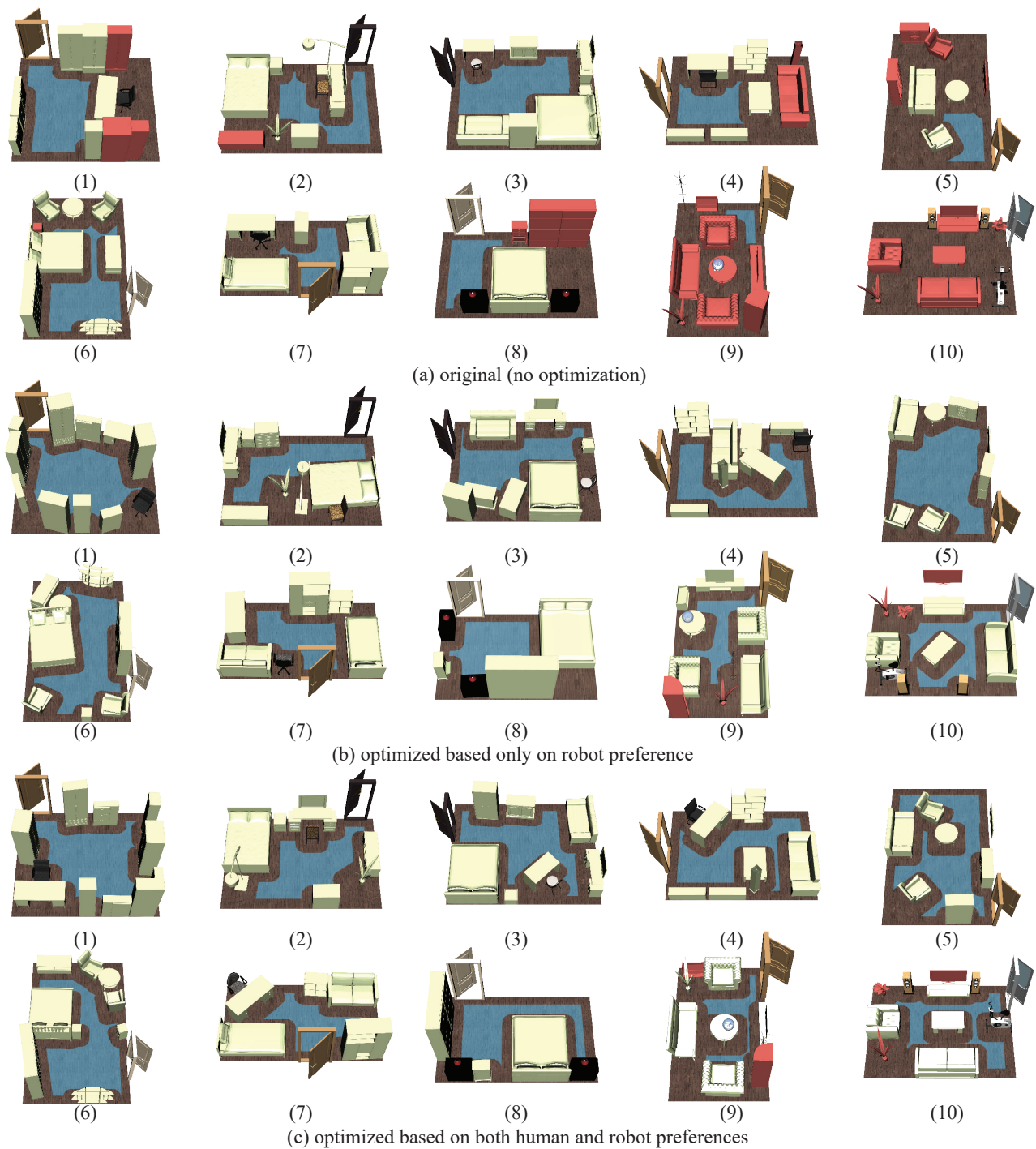}
    \caption{\textbf{Examples of rearranged scenes.} \textcolor{CadetBlue}{Blue} area denotes the accessible space, whereas the \textcolor{RedOrange}{red} objects are out of reach.}
    \label{fig:results}
\end{figure*}

\section{Experiments}\label{sec:result}

We test our method on SUNCG~\cite{song2017semantic}. We randomly select 90\% of the scenes for learning and the remaining for testing. Seven scenes were chosen for human evaluation.

\subsection{Qualitative Results}

\cref{fig:results} qualitatively show 10 scenes. The blue shaded area depicts the robot's accessible space, whereas the red objects are out of reach. Comparing optimization based solely on robot preference (\cref{fig:results}b) with optimization based on both human and robot preferences (\cref{fig:results}c), we qualitatively demonstrate that our method (i) simultaneously increases the robot's accessible space and affords more robot interactions with objects, and (ii) properly maintain the human preference encoded by the functional objects; \eg, the desk--chair and bed--nightstand functional object pairs are together.

\subsection{Quantitative Results}

\paragraph*{Robot Preference}

\begin{figure}[t!]
    \centering
    \includegraphics[width=\linewidth]{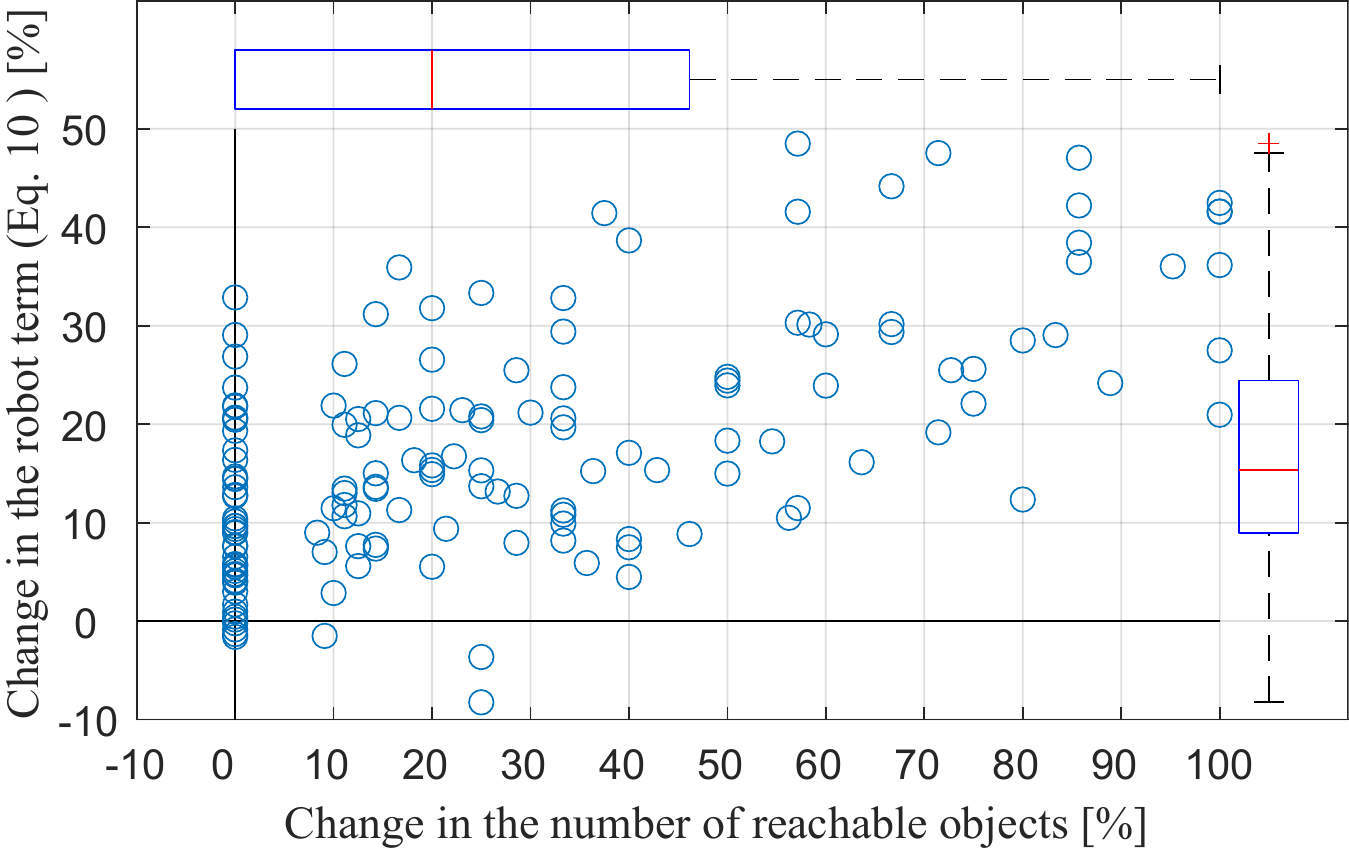}
    \caption{\textbf{Quantitative results evaluated on 154 scenes in SUNCG dataset.} The scatter plot depicts the improvement of each scene in terms of robot preference following scene rearrangement. The boxplots depict the distribution of improvements along each axis, with the red horizontal lines representing the median improvements, and the bottom and top edges of the blue boxes representing the 25th and 75th percentiles, respectively.}
    \label{fig:suncg_result}
\end{figure}

We devise two evaluation criteria: (i) a simple heuristic based on the number of reachable objects, and (ii) the robot term defined in \cref{eqn:robot_loss}.
\cref{fig:suncg_result} summarizes quantitative results aggregated from 154 scenes randomly selected from the test set. Specifically, it plots the change of these two criteria. We note that most of the rearranged scenes are located in the first quadrant, indicating greater support for robot activity. Some scenes are located in the fourth quadrant because interaction spaces were sacrificed in exchange for more accessible objects.

\paragraph*{Human Preference}

We conducted a human study to validate if the rearranged scenes preserve human preference. 11 participants were recruited to evaluate 7 pairs (\cref{fig:results}(1)--(7)) of original (\cref{fig:results}a) and rearranged (\cref{fig:results}c) scenes:
\begin{itemize}[noitemsep,nolistsep,topsep=0pt]
    \item The \textit{Naturalness} (How natural does the scene look?);
    \item The \textit{Functionality} (How well do you think you can perform your daily activities in the scene?).
\end{itemize}
They were asked to provide ratings to the above questions on a scale of 1 to 5, with 1 being \textit{not at all} and 5 \textit{very much}.

The collected responses were analyzed using an independent samples t-test with a significance level of 0.05. Five of seven scenes (\ie, \cref{fig:results}(1)--(5)) lack statistical significance for either \textit{Naturalness} or \textit{Functionality}. Although the rearranged scenes tend to obtain slightly lower human ratings than their original layouts, the insignificance indicates that the rearranged scenes do not significantly diminish human values, while improving the robots' accessible space and the number of supported activities.

\begin{figure}[t!]
    \centering
    \includegraphics[width=\linewidth]{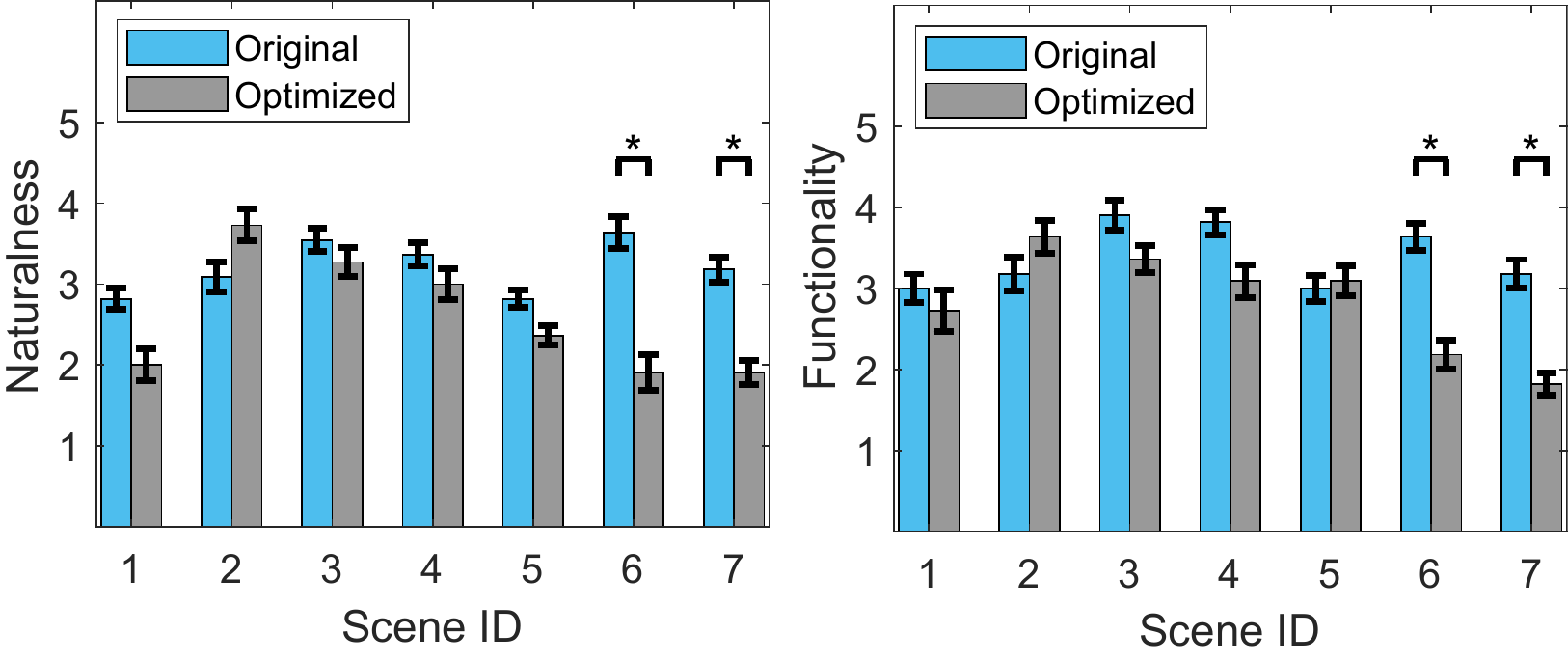}
    \caption{\textbf{Averaged human rating scores regarding whether or not rearranged scenes maintain preferences in terms of \textit{Naturalness} and \textit{Functionality}.} The black error bars represent a 95\% confidence interval. The first five pairs of scenes following the order in \cref{fig:results} are not statistically significant, whereas the last two are.}
    \label{fig:human_result}
\end{figure}  

The differences in human rating for scene 6 and scene 7 are statistically significant, with $t(10) = 1.0, p = 0.0029$ and $t(10) = 1.0, p = 0.0023$, respectively. Although the rearranged scenes are more friendly to robots, human participants appear to be quite sensitive to unusual object arrangements. After rearrangement, they can quickly identify the undesired orientation of the bed in \cref{fig:results}(6) and the chair blocked by the desk in \cref{fig:results}(7), resulting in much lower ratings for these two scenes.

\subsection{Ablations}

\paragraph*{Robot Factors}

\begin{figure}[t!]
    \centering
    \includegraphics[width=\linewidth]{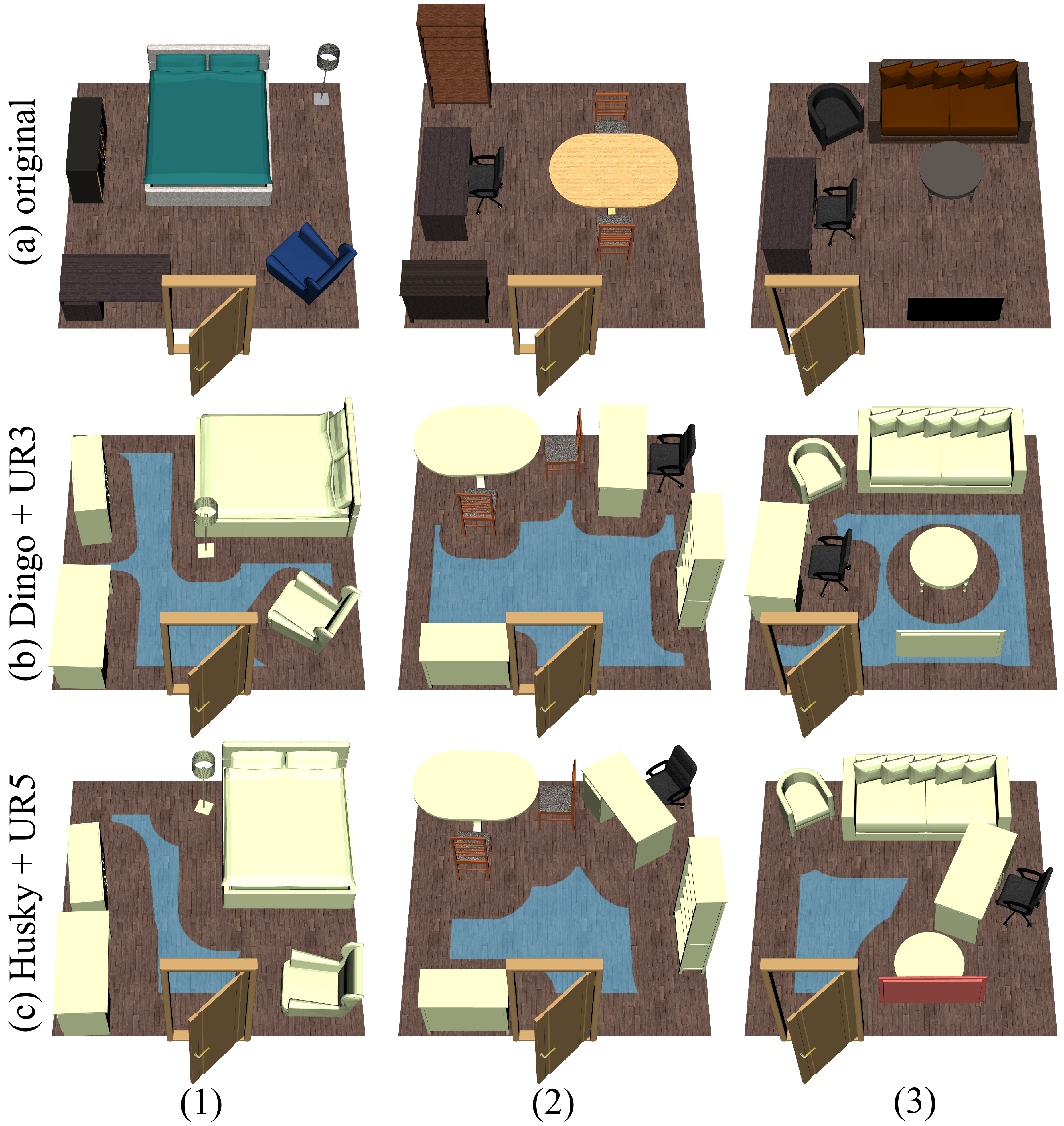}
    \caption{\textbf{Scenes rearranged for Dingo and Husky.}}
    \label{fig:exp_ablation_robot}
\end{figure}

To further study how robot specifications may impact the results of scene rearrangement, we use three living areas presented in \cref{fig:exp_ablation_robot}a to illustrate how our method performs differently for two distinct robots: Husky with UR5, which is larger, and Dingo with UR3, which is smaller. 
The scenes in \cref{fig:exp_ablation_robot}(1) and \cref{fig:exp_ablation_robot}(2) depict the scenes optimized for Dingo and Husky robots, respectively. Despite the final layouts for both robots (\cref{fig:exp_ablation_robot}b and \cref{fig:exp_ablation_robot}c) are similar, the layout in \cref{fig:exp_ablation_robot}(3) is still not optimal for the Husky robot due to its larger size. These results suggest that if the robot's dimension exceeds a particular threshold, it may be impossible to design a layout that accommodates the robot's activities. Similarly, the improvements offered by the scene rearrangement based on a specific type of robot are only partially transferable to another, possibly requiring an entirely new scene rearrangement.

\paragraph*{Task-Specific Factors}

\begin{figure}[t!]
    \centering
    \includegraphics[width=\linewidth]{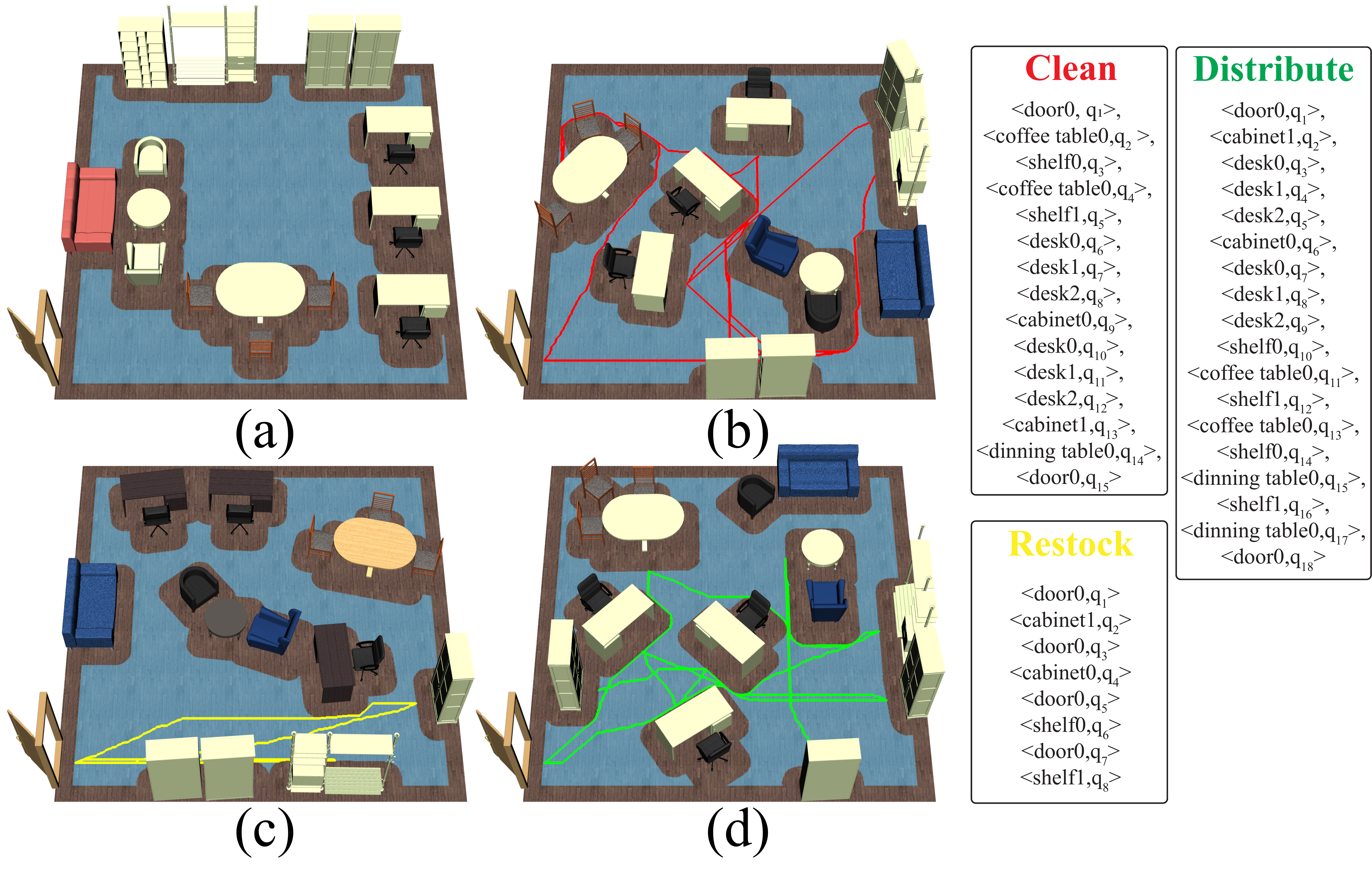}
    \caption{\textbf{Scenes optimized for human-robot co-activity given various activities.}}
    \label{fig:exp_ablation_task}
\end{figure}

In this ablation, we demonstrate the applicability of our method to task-specific scenarios by defining \textit{Robot Motion Cost} $M_T = \sum_{t \in T} \mathcal{L} ( p(q^t_i, q^t_j) ) $ as the robot's total navigation distance of a set of activities $T$. Adding this cost term to the optimization objective defined in \cref{eqn:objective} can rearrange scenes to reduce the robot's motion effort further, hence improving its task efficiency. In an office environment depicted in \cref{fig:exp_ablation_task}a, which contains three types of functional groups---working, relaxing, and dining---the robot has been assigned three tasks involving numerous activities. The ``clean'' task (red paths, \cref{fig:exp_ablation_task}b) is performed primarily in the working functional group, the ``restocking'' task (yellow paths, \cref{fig:exp_ablation_task}c) requires the robot to visit all cabinets and shelves in the scene, and the ``distribution'' task (green paths, \cref{fig:exp_ablation_task}d) requires the robot to traverse among all three functional groups. These scenes are rearranged according to the Dingo robot's specifications.

\section{Conclusion}\label{sec:conclude}

This paper presented an optimization-based framework for rearranging indoor scenes according to both human and robot preferences. Specifically, the human preference was captured by uncovering the functional relations among objects governing their arrangements, modeled by their semantic and spatial co-occurrences from the ConceptNet and SUNCG datasets. The robot preference was represented by its accessible space. We formulated these factors into an optimization problem that rearranges a given scene by optimizing furniture layouts. Experimental results showed that the proposed method expands the open space, increases the number of reachable objects, and minimizes traveling effort in robot activities. Moreover, our human study revealed that most of the rearranged scenes remained natural and acceptable to humans, as the ratings were statistically insignificant compared to the original layouts. These findings signified that rearranges produced by the proposed method effectively promote human-robot co-activities.

\paragraph*{Acknowledgments}
This work is supported in part by the National Key R\&D Program of China (2022ZD0114900), the Beijing Municipal Science \& Technology Commission (Z221100003422004), and the Beijing Nova Program.

\balance
\bibliographystyle{ieeetr}
\bibliography{reference}
\end{document}